\documentclass[10pt,twocolumn,letterpaper]{article}
\usepackage[pagenumbers]{cvpr} 

\usepackage[dvipsnames]{xcolor}

\definecolor{cvprblue}{rgb}{0.21,0.49,0.74}
\usepackage[pagebackref,breaklinks,colorlinks,citecolor=cvprblue]{hyperref}

\usepackage{algorithm, algorithmic}

\usepackage{etoolbox}
\usepackage{graphicx}
\usepackage{amsmath}
\usepackage{amssymb}
\usepackage{booktabs}
\usepackage{makecell}
\usepackage{multirow}
\usepackage{overpic}
\usepackage{comment}
\usepackage{float}
\usepackage[export]{adjustbox}
\usepackage{marvosym}

\usepackage[capitalize]{cleveref}

\newcommand\sotaa{\textcolor{red}}
\newcommand\sotab{\textcolor{blue}}

\def\vs{\emph{vs.\ }}
\def\eg{\emph{e.g.,\ }}
\def\ie{\emph{i.e.,\ }}

\def\etal{\emph{et al.\ }}

\def\paperID{xxx} 
\def\confName{CVPR}
\def\confYear{2024}

\begin{document}

\title{Key-Graph Transformer for Image Restoration}

\newcommand*\samethanks[1][\value{footnote}]{\footnotemark[#1]}
\author{
Bin Ren$^{1,2}$\thanks{Equal contribution}, Yawei Li$^{3}$\samethanks, Jingyun Liang$^{3}$, Rakesh Ranjan$^{4}$,\\Mengyuan Liu$^{5}$, Rita Cucchiara$^{6}$, Luc Van Gool$^{3}$, Nicu Sebe$^{2}$ \\
$^1$University of Pisa \quad
$^2$University of Trento \quad 
$^3$ETH Z\"urich \\
$^4$Meta Reality Labs,
$^5$Peking University,
$^6$University of Modena and Reggio Emilia\\
}

\maketitle

\begin{abstract}
While it is crucial to capture global information for effective image restoration (IR), integrating such cues into transformer-based methods becomes computationally expensive, especially with high input resolution. Furthermore, the self-attention mechanism in transformers is prone to considering unnecessary global cues from unrelated objects or regions, introducing computational inefficiencies. 
In response to these challenges, we introduce the Key-Graph Transformer (KGT) in this paper. 
Specifically, KGT views patch features as graph nodes. The proposed Key-Graph Constructor efficiently forms a sparse yet representative Key-Graph by selectively connecting essential nodes instead of all the nodes.
Then the proposed Key-Graph Attention is conducted under the guidance of the Key-Graph only among selected nodes with linear computational complexity within each window.
Extensive experiments across 6 IR tasks confirm the proposed KGT's state-of-the-art performance, showcasing advancements both quantitatively and qualitatively. 
\end{abstract}

\section{Introduction}
\label{sec:introduction}
Image restoration (IR), a fundamental task in the realm of low-level computer vision, is dedicated to the quality improvement of images that have been compromised by various factors such as noise, blur, low resolution, compression artifact, mosaic, adverse weather, or other forms of distortion. 
This capability finds diverse applications, including information recovery (such as retrieving obscured data in medical imaging, surveillance, and satellite imagery) and supporting downstream vision tasks like object detection, recognition, and tracking~\cite{sezan1982image,molina2001image}. Despite significant advancements in recent years, it is noteworthy that current popular image restoration methods still face challenges in effectively handling complex distortions or preserving/recovering essential image details~\cite{li2023efficient}. To recover high-quality images, the rich information exhibited in the degraded counterparts needs to be exquisitely explored.

\begin{figure}[!t]
    \centering
    \includegraphics[width=1.0\linewidth]{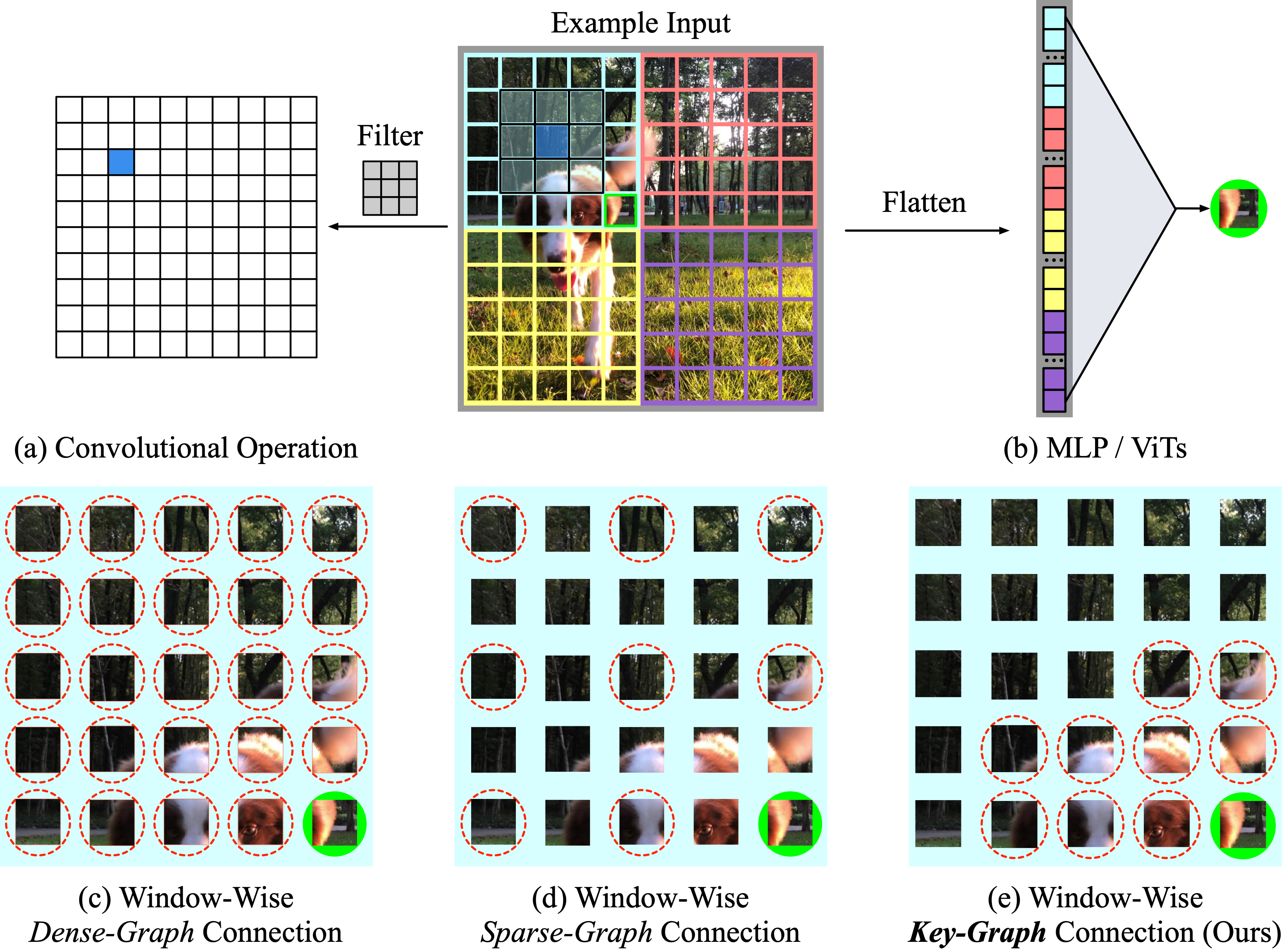}
    \caption{(a) The CNN filter captures information only within a local region. (b) The standard MLP/Transformer architectures take full input in a long sequence manner. (c) The window-size multi-head self-attention (MSA) mechanism builds a fully connected dense graph within each window. (d) Position-fixed sparse graph. (e) The proposed Key-Graph connects only the essential nodes.}
    \label{fig:dog_demo}
    \vspace{-3mm}
\end{figure}

In modern computer vision systems, representative networks for learning rich image information in IR are primarily constructed using 3 fundamental architectural paradigms. \ie the convolutional neural networks (CNNs)~\cite{lecun1998gradient,zamir2021multi}, Multilayer perceptrons (MLPs)~\cite{bishop2006pattern,tu2022maxim}, and Vision Transformers (ViTs)~\cite{vaswani2017attention,dosovitskiy2020image}. The input image is treated as a regular grid of pixels in the Euclidean space for CNNs (Fig.~\ref{fig:dog_demo}(a)) or a sequence of patches for MLPs and ViTs (Fig.~\ref{fig:dog_demo}(b)). However, the degraded input usually contains irregular and complex objects. These choices perform admirably in specific scenarios characterized by regular or well-organized object boundaries but have limitations when applied to images with more flexible and complex geometrical contexts.

Additionally, CNNs are struggling to model the long-range dependencies because of their limited receptive field (Fig.~\ref{fig:dog_demo}(a)). MLPs/ViTs are widely validated for capturing the long-range relation but at the cost of losing the ability for inductive bias or heavy computation burden (\eg quadratic complexity increases with the increase of the input resolution)~\cite{tu2022maxim,vaswani2017attention,dosovitskiy2020image,ren2023masked}.). 
To overcome these limitations, recent methods investigate strategies for complexity reduction. One common approach is to implement MSA within local image regions~\cite{liu2021swin}, \eg a full MSA or a region-fixed anchored stripe MSA is conducted by SwinIR~\cite{liang2021swinir} or GRL~\cite{li2023efficient}, which still struggles to capture inherent connections among irregular objects. Additionally, an earlier study~\cite{zontak2011internal} highlights that smooth image contents occur more frequently than complex image details, suggesting the need for differentiated treatment for different contents.

In this paper, we introduce a novel approach, the Key-Graph Transformer (KGT), to address the limitations above. Our method comprises two core components: the k-nearest neighbors (KNN) based \textit{Key-Graph Constructor} and a \textit{Key-Graph Transformer layer} integrated with a novel Key-Graph attention block. Specifically, starting with the low-level feature obtained from the convolutional feature extractor, each patch is treated as a node of a graph. Since capturing long-range dependencies among all nodes (Fig.~\ref{fig:dog_demo}(c)) can be highly computationally demanding, we selectively choose $k$ essential nodes (Fig.~\ref{fig:dog_demo}(e)) based on the proposed Key-Graph constructor rather than establishing connections in a sparse yet position-fixed manner. (Fig.~\ref{fig:dog_demo}(d)). This leads to a sparse yet representative graph that connects only the essential nodes, which allows our method to achieve the same receptive field as previous ViTs-based methods while maintaining lower computational costs.
The criteria for selecting these nodes are determined by the self-similarity calculated at the beginning of each KGT layer. 
Then the chosen nodes undergo processing by all the successive Key-Graph transformer layers.
It's worth noting that the implementation of the Key-Graph attention block within each KGT layer is achieved in three interesting manners (\ie the Triton~\cite{dao2022flashattention}\footnote{Open-source GPU programming tool \url{https://openai.com/research/triton}.}, torch-mask, and torch-gather), which will be discussed in our ablation studies.
Based on these two components, together with a convolutional operation at the end of each KGT stage, the global information that exists in all the selected nodes is well-aggregated and updated. 

In summary, our main contributions are listed as follows:
\begin{enumerate}[nosep]
    \item We propose a Key-Graph constructor that provides a sparse yet representative \textit{Key-Graph} with the most relevant $k$ nodes considered, which works as a reference for the subsequent attention layer, facilitating more efficient attention operations.

    \item Based on the constructed Key-Graph, we introduce a Key-Graph Transformer layer with a novel Key-Graph attention block integrated. Notably, the computational complexity can be significantly reduced compared to conventional attention operations.
    
    \item We propose the KGT for IR. Extensive experimental results show that the proposed KGT achieves state-of-the-art performance on 6 IR tasks, \ie deblurring, JPEG compression artifact removal (JPEG CAR), denoising, IR in adverse weather conditions (AWC), demosaicking, and classic image super-resolution (SR).
\end{enumerate}
\section{Related Work}
\label{sec:related-work}
\noindent{\textbf{Image Restoration (IR),}} as a long-standing ill-posed inverse problem, is designed to reconstruct the high-quality image from the corresponding degraded counterpart. It has been brought to various real-life scenarios due to its valuable application property~\cite{richardson1972bayesian,banham1997digital,li2023lsdir}. Initially, IR was addressed through model-based solutions, involving the search for solutions to specific formulations. However, with the remarkable advancements in deep neural networks, learning-based approaches have gained increasing popularity. These approaches have been explored from various angles, encompassing both regression-based~\cite{lim2017enhanced,liang2021swinir,chen2021learning,li2023efficient} and generative model-based pipelines~\cite{gao2023implicit,wang2022zero,luo2023image,yue2023resshift}. Our focus in this work is to investigate IR under the former pipeline. 

\noindent{\textbf{Non-Local Priors Modeling in IR.}} Tradition model-based IR methods reconstruct the image by regularizing the results (\emph{e.g.}, Tikhonov regularization~\cite{golub1999tikhonov}) with formulaic prior knowledge of natural image distribution. However, it's challenging for these model-based methods to recover realistic detailed results with hand-designed priors. Besides, some other classic method finds that self-similarity is an effective prior which leads to an impressive performance~\cite{buades2005non,dabov2007image}. Apart from the traditional methods, the non-local prior also has been utilized in modern deep learning networks~\cite{liu2018non,wang2018non,li2023efficient,zhang2019residual}, and it was usually captured by the self-attention mechanism. Especially, KiT~\cite{lee2022knn} proposed to increase the non-local connectivity between patches of different positions via a KNN matching to better capture the non-local relations between the base patch and other patches in every attention operation, this brings huge extra computation costs. DRSformer~\cite{chen2023learning} proposed a topk selection strategy that chooses the most relevant tokens to model the non-local priors for draining after each self-attention operation without reducing the computation complexity. We aim to further improve the effectiveness of the non-local priors from a more efficient graph perspective.

\noindent{\textbf{Graph-Perspective Solutions for IR.}} Graph is usually used to deal with irregular data structures such as point clouds~\cite{wang2019dynamic,li2021towards}, social networks~\cite{myers2014information}, or protein~\cite{ingraham2019generative}. Recently, it was adapted to process the images in a more flexible manner~\cite{gori2005new,scarselli2008graph,mou2021dynamic,han2022vision,jiang2023graph} on various IR tasks, like facial expression restoration~\cite{liu2020facial}, image denoising~\cite{simonovsky2017dynamic}, and artifact reduction~\cite{mou2021dynamic}. However, most of these solutions for IR mainly extend from graph neural networks (GNNs), which mainly focus on very close neighbor nodes. Merely increasing the depth or width of GNNs proves inadequate for expanding receptive fields~\cite{xu2018representation}, as larger GNNs often face optimization challenges like vanishing gradients and over-smoothing problems. \cite{jiang2023graph} construct the graph with transformer-based architecture but in a very expensive manner where each node is connected to all other nodes. In this paper, we integrate graph properties into ViTs by employing a \textit{Key-Graph} for efficient capture of effective non-local priors for IR.
\begin{figure*}[!t]
    \centering
    \includegraphics[width=1.0\linewidth]{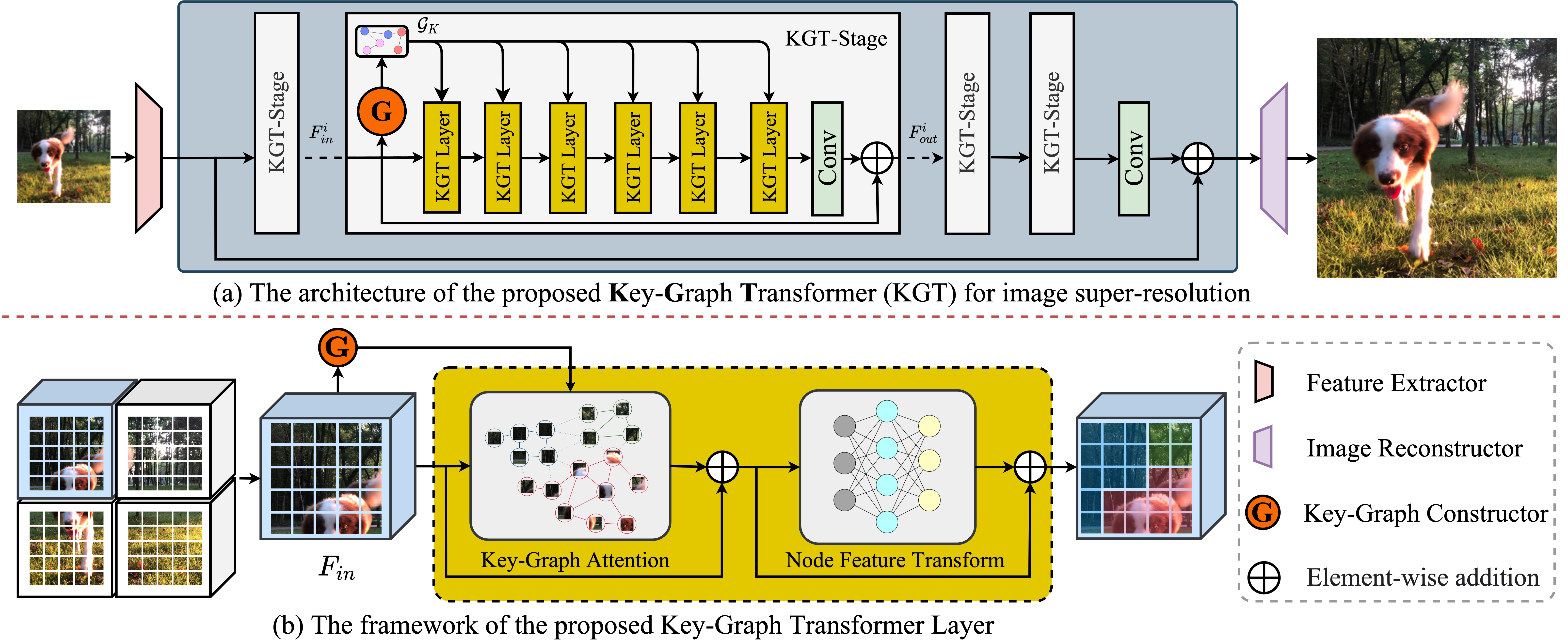}
    \caption{The proposed KGT mainly consists of a convolutional feature extractor, the main body of the proposed KGT for representation learning, and an image reconstructor. The main body shown here is for SR, while the U-shaped structure (Shown in \textit{Appx.}) is used for other IR tasks. (b) The illustration of the Key-Graph Transformer layer within each KGT stage.}
    \vspace{-1em}
    \label{fig:framework}
\end{figure*}
\section{Preliminary: Graph Transformer}
\label{sec:preliminaries}
Graph Transformers generalize Transformers to graphs with all nodes fully connected. Specifically, given input feature $F_{in} \in \mathbb{R}^{H \times W \times C}$, where $H$, $W$, and $C$ denote the height, the width, and the channel, respectively. $F_{in}$ is split into $N$ patches, and the graph nodes are typically assigned based on these patches, forming an unordered node representation $\mathcal{V} = \{v_{i}| v_{i} \in \mathbb{R}^{ hw \times c}, i = 1, 2, 3, ..., N\}$, where $h$, $w$, and $c$ are the height, the width, and the channel of each node. A weighted graph  $\mathcal{G} = (\mathcal{V, E})$, usually described by the weighted adjacency matrix $\mathcal{A}$, is constructed by adding an edge $e_{ji}$ from $v_j$ to $v_i$ among all the neighbors of $v_i$ in $\mathcal{V}$. 

To get $\mathcal{A}$, $\mathcal{V}$ is linearly projected
into Query ($Q$), Key ($K$), and Value ($V$) matrices ($V$ will be used to conduct the node aggregation with the help of $\mathcal{A}$ later), which are denoted as $Q=\mathcal{V}\mathbf{W}_{qry}$, $K=\mathcal{V}\mathbf{W}_{key}$, and $V=\mathcal{V}\mathbf{W}_{val}$. $\mathbf{W}_{qry/key/val}$ represents the learnable projection weights. Then $\mathcal{A}$ is performed by a softmax function as follows:
\begin{equation}
    \mathcal{A}_{ij} = \frac{\operatorname{exp}(Q_{i}K_{j}^{T})}{\sum_{k = 1...j} \operatorname{exp}(Q_{i}K_{k}^{T}/\sqrt{d})}, 
    \label{eq:eq1}
\end{equation}
where $d$ represents the dimension of $Q$ and $K$. Then the node feature can be aggregated to $\hat{v_i}$ by:
\begin{equation}
 \hat{v_i} = \mathcal{A}_{ij} V_{i} = \mathcal{A}_{ij}\mathcal{V}\mathbf{W}_{val}.
 \label{eq:eq2}
\end{equation}
However, $\mathcal{A}_{ij}$ in standard ViTs describes a fully connected graph, \eg given a sub-graph with a green dog root node shown in Fig.~\ref{fig:dog_demo}(c), the tree-related nodes are also considered. Such dense connection largely limits the efficiency of ViTs on large-scale input. To mitigate this problem, we assumed that for each node, a sub-graph with only the necessary connection is sufficient to find the balance between performance and efficiency. To this end, we aim to achieve a sparse yet representative adjacency matrix $\mathcal{A}_{K}$ which describes a flexible and sparse graph $\mathcal{G}_{K}$ ( Fig.~\ref{fig:dog_demo}(e)) that connect the essential nodes for a destination node.

\begin{algorithm}[!t]
    \small
    \renewcommand{\algorithmicrequire}{\textbf{Input:}}
    \renewcommand{\algorithmicensure}{\textbf{Output:}}
    \caption{Key-Graph Transformer Stage}
    \label{alg:kgt_stage}
    \begin{algorithmic}[1]
        \REQUIRE input feature ${F_{in}}$, numbers of KGT layer $N_{layer}$, KNN value $k$, the patched node feature $\mathcal{V}$
	\ENSURE aggregated feature $F_{out}$
            \STATE $\mathcal{G}_{K} \leftarrow \operatorname{KeyGraph\_Constructor}(\mathcal{V}, k)$ // Sec.~\ref{subsec:KG_cons}
            \FOR{$i=1$ to $N_{layer}$}
                \STATE $Q, K, V \leftarrow \operatorname{Linear\_Proj}(\mathcal{V})$ 
                \STATE $\hat{\mathcal{V}} \leftarrow \operatorname{KeyGraph\_Att}(Q, K, \mathcal{G}_{K})$ // Sec.~\ref{subsec:kgt_layer}
                \STATE $\mathcal{Z} \leftarrow \hat{\mathcal{V}} + \operatorname{FFN}(\hat{\mathcal{V}})$ // Sec.~\ref{subsec:kgt_layer}
            \ENDFOR
            \STATE $F_{out} \leftarrow F_{in} + \operatorname{Conv}(\mathcal{Z})$ 
		\STATE \textbf{return} $F_{out}$
    \end{algorithmic}  
\end{algorithm}

We adopted the window-wise MSA throughout our method, to streamline our explanation, we select a single window for illustration when discussing the proposed Key-Graph Constructor and Key-Graph Transformer layer. Notations such as $F_{in}$ and $\mathcal{V}$ are also window-size adapted for clarity.

\section{Methodology}
\label{sec:method}
Unlike conventional approaches that treat $F_{in}$ after the feature extractor as a regular grid of pixels (typical in CNNs) or as a sequence of patches (common in MLPs and ViTs), we adopt a flexible graph representation manner. The overall architecture of the proposed Key-Graph Transformer (KGT) is shown in Fig.~\ref{fig:framework}, and we formalize the pipeline of each KGT stage in Alg.~\ref{alg:kgt_stage}. Each step will be introduced in the corresponding subsections in detail. Specifically, at the beginning of each KGT stage, a sparse yet representative graph $\mathcal{G}_{K}$ will be constructed. The efficiency of graph updating is ensured by the Key-Graph Constructor (Sec.~\ref{subsec:KG_cons}) in a shared manner within each stage. Simultaneously, the effectiveness is achieved by the Key-Graph Transformer Layer (Sec.~\ref{subsec:kgt_layer}). $\operatorname{Conv()}$ is applied together with a residual connection as the last step of Alg.~\ref{alg:kgt_stage}. Two interesting discussions (Sec.\ref{subsec:discuss}) are introduced regarding the implementation style of the Key-Graph attention and two topk settings during the training.

\subsection{Key-Graph Constructor}
\label{subsec:KG_cons}
The proposed Key-Graph constructor aims to construct a sparse yet representative graph $\mathcal{G}_{K}$ which will be used for conducting the efficient Key-Graph attention operation for all the following KGT layers within the same stage in a shared manner. Specifically, given $\mathcal{V}$, an initial fully connected graph $\mathcal{G}$ is constructed by calculating the self-similarity $\operatorname{Sim()}$ of $\mathcal{V}$ via naive dot product operation and outputs the corresponding adjacency matrix $\mathcal{A}$ as below:
\begin{equation}
\mathcal{A} = \operatorname{Sim}(i, j ) = v_i \cdot v_j^{T},
\end{equation}
which describes the correlation among all the nodes. A higher value indicates a higher correlation. However, in this context, $\mathcal{A}$ represents a fully connected dense graph, wherein all nodes $v_j$ within $\mathcal{V}$ are included in the connectivity of the destination node $v_i$, irrespective of the degree of semantic relatedness between $v_i$ and $v_j$.

To mitigate the side effects of nodes with low correlation (\eg the tree-related nodes at the upper left part in Fig.~\ref{fig:dog_demo} (c)) for the green background dog destination node, we keep only $k$ highly related nodes of the destination node $v_i$ and exclude the remainings.
This is achieved by a KNN algorithm from $\mathcal{A}$ as follows:
\begin{equation}
\mathcal{A}_{K}(i, j) = 
    \begin{cases}
    \mathcal{A}(i, j),~\mathcal{A}(i, j) \geq \operatorname{Sim}(i, )_{k} ~and~ i \neq j
    \\ 
    0,~~~~~~~~~~otherwise,
    \end{cases}
\end{equation}
where $\operatorname{Sim}(i, )_{k}$ denotes the $k_{th}$ largest connective value of node $v_{i}$ with the corresponding node. As a result, $\mathcal{G}_{K}$ is achieved which contains only the nodes with high correlation (\eg dog-related nodes in Fig.~\ref{fig:dog_demo}(e)) for the destination node (\eg the green dog node). We formalize the graph constructor process as $\operatorname{KeyGraph\_Constructor}()$ in Alg.~\ref{alg:kgt_stage}.

Owing to the permutation-invariant property inherent in both the MSA and the FFN within each transformer layer~\cite{vaswani2017attention,lee2019set}, the KGT layer consistently produces identical representations for nodes that share the same attributes, regardless of their positions or the surrounding structures~\cite{chen2022structure}. In other words, nodes at the same location are consistently connected to other nodes possessing the same attributes as they traverse through the various layers within the same stage. This enables $\mathcal{G}_{K}$ as a reference for each attention block in the subsequent KGT layers within each stage, facilitating efficient attention operations. This is different from the sparse graph (See Fig.~\ref{fig:dog_demo}(d)) that only activates the nodes in a fixed coordinate of a given feature~\cite{zhang2023accurate}.

\subsection{Key-Graph Transformer Layer}
\label{subsec:kgt_layer}
The proposed Key-Graph Transformer Layer is shown in Fig.~\ref{fig:framework}(b), which mainly consists of a Key-Graph attention block followed by a feed-forward network (FFN).

\begin{figure}[!t]
    \centering
    \includegraphics[width=1.0\linewidth]{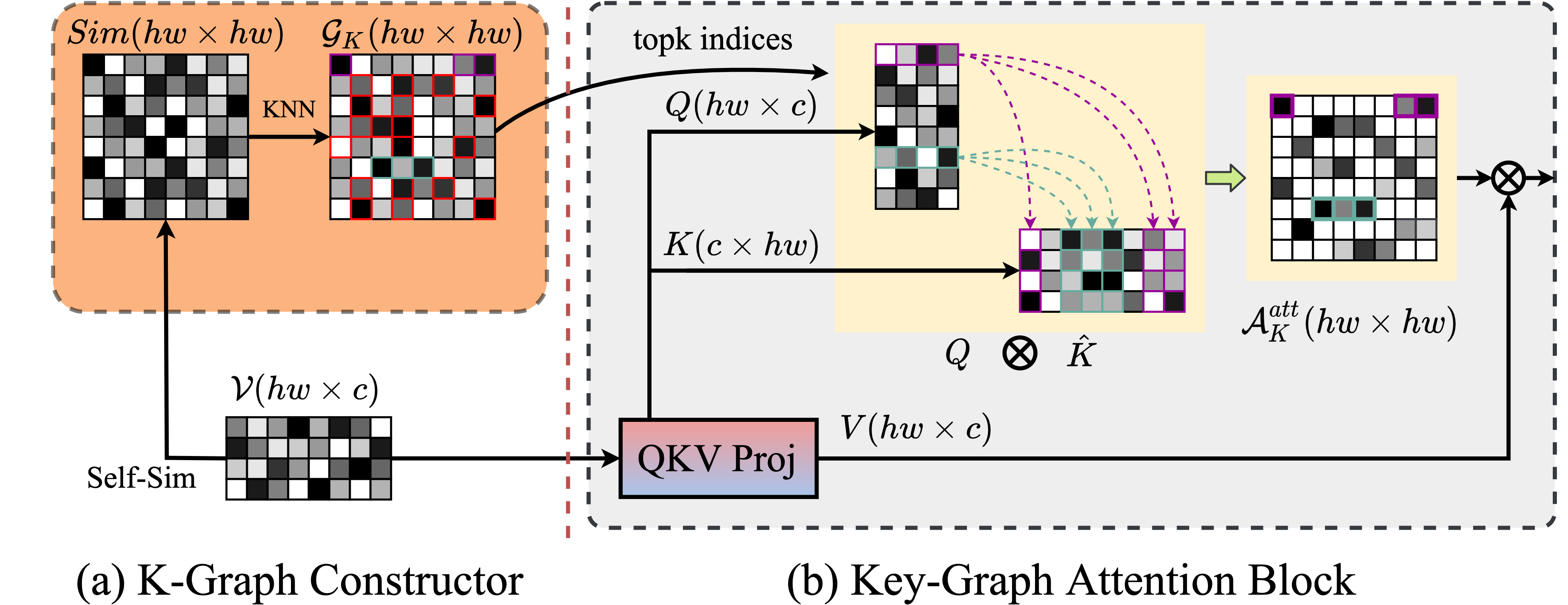}
    \caption{The toy example of $k$=3 for the illustration of Key-Graph Constructor (a) and the Key-Graph attention (b) within each KGT Layer.}
    \label{fig:k_graph}
    \vspace{-3mm}
\end{figure}

Fig.~\ref{fig:k_graph}(b) shows the detailed workflow of the proposed Key-Graph attention block. 
Initially, the node $\mathcal{V}$ is linear projected via $\operatorname{Linear\_Proj()}$ (The 3rd step in Alg.~\ref{alg:kgt_stage}) into $Q$, $K$, and $V$.
Then for each node $v_i$ in $Q$, instead of calculating the self-attention with all the $hw$ nodes in $K$, only $k$ essential nodes in $K$ are selected under the guidance of $\mathcal{G}_{K}$, forming the $\hat{K}$. 
We intuitively show such a process in Fig.~\ref{fig:k_graph} (a) and (b). Then the current spare yet representative adjacency matrix $\mathcal{A}_{K}$ is obtained by:
\begin{equation}
    \mathcal{A}_{K}^{att} = \operatorname{Softmax_{K}}(Q\hat{K}^{T}/ \sqrt{d}),
    \label{eq:atten}
\end{equation}
which captures the pair-wise relation between each destination node $v_i$ in $Q$ with only the $k$ nodes in $K$ that are semantically highly related to $v_i$ in the current KGT layer. For other nodes apart from the selected $k$ nodes, our idea is to keep their position in their corresponding places without any computation. Based on $\mathcal{A}_{K}^{att}$, the Key-Graph attention outputs the updated node feature $\hat{\mathcal{V}}$ via Eq.~\ref{eq:eq2}. This is different from the conventional MSA which calculates the relation of each node in $Q$ and all nodes in $K$ (See the difference between (c) \& (e) in Fig.~\ref{fig:dog_demo}). Meanwhile, our Key-Graph attention is also different from the sparse attention~\cite{zhang2023accurate} where the nodes that need to be collected are always in a fixed position (See the difference between (d) \& (e) in Fig.~\ref{fig:dog_demo}). Conversely, our Key-Graph attention block not only significantly reduces the computational complexity from $\mathcal{O}((hw)^{2})$ to $\mathcal{O}((hw) \times k)$ within each window, where $k < hw$, but also provides a more flexible approach to capturing semantically highly related nodes.

Finally, as over-smoothing is prevalent in graph-structured data, it becomes particularly pronounced in deep models~\cite{chen2020measuring,keriven2022not} even for Transformers~\cite{zhai2023stabilizing,noci2022signal}. 
To relieve such loss of distinctive representation~\cite{oono2019graph} and encourage the node feature transformation capacity, we adopted the FFN on each node feature together with a residual connection operation. 
This process can be formalized as follows:
\begin{equation}
\mathcal{Z} = \hat{\mathcal{V}} + \operatorname{FFN}(\hat{\mathcal{V}}) = \hat{\mathcal{V}} + \sigma(\mathcal{\hat{V}}\mathbf{W_{1}})\mathbf{W_{2}},
\end{equation}  
where $\mathcal{Z} \in \mathbb{R}^{hw \times c}$ is the transformed node feature. $\sigma$ is the activation function. $\mathbf{W_{1}}$ and $\mathbf{W_{2}}$ are the learnable weights of two MLPs in $\operatorname{FFN()}$.

\subsection{Discussion}
\label{subsec:discuss}
\noindent \textbf{Implementation of Key-Graph Attention.} To achieve the proposed Key-Graph attention operation,  
we explored three different manners for the detailed implementation, \ie 
(i) \textit{Triton}, (ii) \textit{Torch-Gather}, and (iii) \textit{Torch-Mask}. Specifically, (i) is based on FlashAttention~\cite{dao2022flashattention}, and a customized GPU kernel is written for the operators proposed in this paper. Parallel GPU kernels are called for the nodes during run time. (ii) means that we use the `torch.gather()' function in PyTorch to choose the corresponding $Q_{gather}$ and $K_{gather}$ based on $\mathcal{G}_{K}$, then the attention operation shown in Eq.~\ref{eq:atten} is conducted between $Q_{gather}$ and $K_{gather}$. (iii) denotes that we keep only the value of selected nodes of $\mathcal{A}_{K}$ and omitting other nodes with low correlation via assigning those values to $-\infty$ guided by $\mathcal{G}_{K}$. We will discuss the pros and cons of these manners in Sec.~\ref{subsec:ablatin}.

\noindent \textbf{Fixed topk \vs Random topk Training Strategies.} For fixed topk strategy, $k$ is fixed to 512 during training. For random topk, $k$ is randomly selected from the values $[64, 128, 192, 256, 384, 512]$ during training. Note that during inference $k$ is configured to the specified value according to the computational budget.
\section{Experiments}
\label{sec:experiments}
\begin{figure}[!t]
    \centering
    \includegraphics[width=1\linewidth]{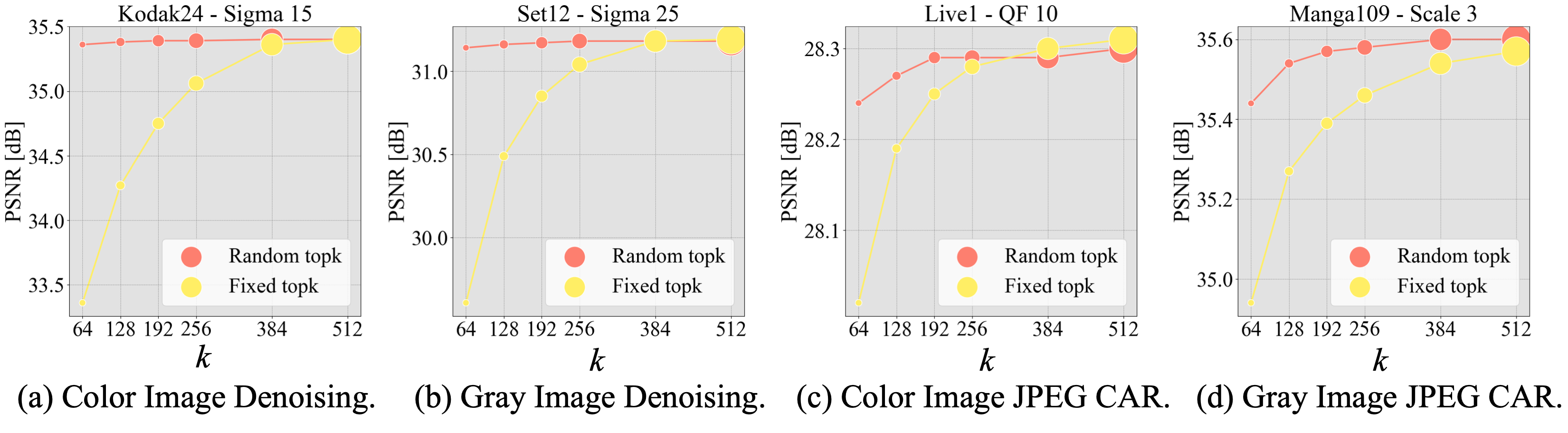}
    \vspace{-3mm}
    \caption{Ablation study on the impact of $k$. 
 The size of the circle denotes the FLOPs. The $k$ on the horizontal axis is the one used during inference.}
    \label{fig:main_figure_K}
    \vspace{-3mm}
\end{figure}

\begin{table*}[!t]
\parbox{.34\linewidth}{
\scriptsize
\begin{center}
\caption{GPU memory footprint of different implementations (\ie Triton, Torch-Gather, and Torch-Mask) of our key-graph attention block. $N$ is the number of tokens and $k$ is the number of nearest neighbors. OOM denotes "out of memory".}
\label{table:ablation_implementation}
\vspace{-0.1mm}
\setlength{\tabcolsep}{7pt}
\begin{tabular}{c|ccc}
\toprule[0.15em]

$N$	&Triton	&Torch-Gather	&Torch-Mask	\\ \midrule
512	&0.27 GB	&0.66 GB	&0.36 GB	\\
1024	&0.33 GB	&1.10 GB	&0.67 GB	\\
2048	&0.68 GB	&2.08 GB	&1.91 GB	\\
4096	&2.61 GB	&4.41 GB	&6.83 GB	\\
8192	&10.21 GB	&10.57 GB	&26.42 GB	\\ \midrule[0.15em]

$k$	&Triton	&Torch-Gather	&Torch-Mask	\\ \midrule
32	&5.51 GB	&15.00 GB	&13.68 GB	\\
64	&5.82 GB	&27.56 GB	&13.93 GB	\\
128	&6.45 GB	&OOM	&14.43 GB	\\
256	&7.70 GB	&OOM	&15.43 GB	\\
512	&10.20 GB	&OOM	&17.43 GB	\\

\bottomrule[0.15em]
\end{tabular}
\end{center}
}
\hfill
\parbox{.63\linewidth}{\scriptsize\centering
\caption{\textit{\textbf{Single-image motion deblurring}} results. {GoPro}~\cite{nah2017deep} dataset is used for training. 
}
\label{table:motion_deblurring}
\vspace{-3mm}
\setlength{\tabcolsep}{8.8pt}
\begin{tabular}{l | cc | cc | cc}
\toprule[0.15em]
 & \multicolumn{2}{c|}{\textbf{GoPro} } & \multicolumn{2}{c|}{\textbf{HIDE} } & \multicolumn{2}{c}{Average} \\
 \textbf{Method} & PSNR$\uparrow$ & {SSIM$\uparrow$} & PSNR$\uparrow$ & {SSIM$\uparrow$} & PSNR$\uparrow$ & {SSIM$\uparrow$} \\
\midrule[0.1em]
DeblurGAN~\cite{deblurgan}	&28.70 & 0.858		&24.51 & 0.871		&26.61 & 0.865		\\
Nah~\etal~\cite{nah2017deep}	&29.08 & 0.914		&25.73 & 0.874		&27.41 & 0.894		\\
DeblurGAN-v2~\cite{deblurganv2}	&29.55 & 0.934		&26.61 & 0.875		&28.08 & 0.905		\\
SRN~\cite{tao2018scale}	&30.26 & 0.934		&28.36 & 0.915		&29.31 & 0.925		\\
Gao \etal \cite{gao2019dynamic}	&30.90 & 0.935		&29.11 & 0.913		&30.01 & 0.924		\\
DBGAN \cite{zhang2020dbgan}	&31.10 & 0.942		&28.94 & 0.915		&30.02 & 0.929		\\
MT-RNN \cite{mtrnn2020}	&31.15 & 0.945		&29.15 & 0.918		&30.15 & 0.932		\\
DMPHN \cite{dmphn2019}	&31.20 & 0.940		&29.09 & 0.924		&30.15 & 0.932		\\
Suin \etal \cite{Maitreya2020}	&31.85 & 0.948		&29.98 & 0.930		&30.92 & 0.939		\\
CODE~\cite{zhao2023comprehensive} & 31.94 & - & 29.67 & - & 30.81  &  -  \\
SPAIR~\cite{purohit2021spatially_spair}	&32.06 & 0.953		&30.29 & 0.931		&31.18 & 0.942		\\
MIMO-UNet+~\cite{cho2021rethinking_mimo}	&32.45 & 0.957		&29.99 & 0.930		&31.22 & 0.944		\\
IPT~\cite{chen2021pre}	&32.52 & -		&- & -		&- & -		\\
MPRNet~\cite{zamir2021multi}	&32.66 & 0.959		&30.96 & 0.939		&31.81 & 0.949		\\
KiT~\cite{lee2022knn} &  32.70 & 0.959 & 30.98 & \sotaa{0.942}  & 31.84 & 0.951  \\  
Restormer~\cite{zamir2022restormer}	&32.92 & 0.961		&\textcolor{red}{31.22} & \textcolor{red}{0.942}	&32.07 & \textcolor{blue}{0.952}
		\\
Ren \etal~\cite{ren2023multiscale} & \sotab{33.20} & \sotab{0.963} & 30.96 & 0.938 & \sotab{32.08} & 0.951  \\
KGT (ours)	&\textcolor{red}{33.44} & \textcolor{red}{0.964}		&\textcolor{blue}{31.05} & \textcolor{blue}{0.941}		&\textcolor{red}{32.25} & \textcolor{red}{0.953}	\\
\bottomrule[0.15em]
\end{tabular}
}
\vspace{-4mm}
\end{table*}
\begin{table}[!t]
    \centering
    \caption{The comparison of different methods regarding the parameters, runtime, and PSNR on Urban100 for $\times4$ SR.}
    \label{tab:ab_efficiency1}
    \setlength{\tabcolsep}{8pt}
    \setlength{\extrarowheight}{0.5pt}
    \scalebox{0.8}{
    \begin{tabular}{cccc}
        \toprule[0.15em]
        Method & Params (M) & Runtime (ms) & PSNR \\ \hline
        SwinIR~\cite{liang2021swinir} & 11.90 & 152.24 & 27.45 \\
        ART~\cite{zhang2023accurate} & 16.55 & 248.26 & 27.77 \\
        CAT~\cite{chen2022cross} & 16.60 & 357.97 & 27.89 \\
        HAT-S~\cite{chen2023activating} & 9.62 & 306.30 & 27.87 \\
        HAT~\cite{chen2023activating} & 20.62 & 368.61 & 28.37 \\
        KGT-S (Ours) & 12.02 & 211.42 & 28.34 \\ 
        \bottomrule[0.15em]
    \end{tabular}
    }
    \vspace{-2mm}
\end{table}

In this section, we first analyze three important ablation studies of the proposed KGT, followed by extensive experiments on \textbf{6} IR tasks, which include image deblurring, JPEG CAR, image denoising, IR in AWC, image demosaicking, and image SR. Note that we adopt two base architectures, \ie the multi-stage one for image SR and the U-shaped one for the rest IR tasks. More details about the architecture design, training protocols, the training/testing dataset, and additional visual results are shown in the \textit{Appendix (Appx.)}. In addition, the best and the second-best quantitative results are reported in \textcolor{red}{red} and \textcolor{blue}{blue}, respectively. Note that \textcolor{magenta}{\textdagger} denotes a single model that is trained to handle multiple degradation levels \ie noise levels, and quality factors.

\subsection{Ablation Study}
\label{subsec:ablatin}

\noindent\textbf{The impact of the implementation of Key-Graph Attention} is assessed in terms of (i) \textit{Triton}, (ii) \textit{Torch-Gather}, and (iii) \textit{Torch-Mask} under different numbers of N (various from 512 to 8192) and K (various from 32 to 512). The results of the GPU memory footprint are shown in Tab.~\ref{table:ablation_implementation}, which indicate that \textit{Torch-Gather} brings no redundant computation while requiring a large memory footprint. Though \textit{Torch-Mask} brings the GPU memory increase, the increment is affordable compared to \textit{Torch-Gather} and also easy to implement. \textit{Triton} largely saves the GPU memory while at the cost of slow inference and difficult implementation for the back-propagation process. To optimize the efficiency of our KGT, we recommend employing \textit{Torch-Mask} during training and \textit{Triton} during inference, striking a balance between the efficiency and the GPU memory requirement.

\noindent\textbf{The Impact of the $k$ in Key-Graph Constructor.} Two interesting phenomena are observed from the results shown in Fig.~\ref{fig:main_figure_K} regarding the two topk training strategies (See Sec.~\ref{subsec:discuss}). (1) The randomly sampled strategy has a very stable and better performance compared to the fixed topk manner especially when the $k$ is fixed to a small number (\emph{i.e.}, 64, 128, 256). (2) The PSNR can largely increase with the increase of $k$ in a fixed manner. We conclude that a random sampled strategy is more general and stable. It can also make the inference process more flexible regarding different computation resources. More ablation results can be found in our \textit{Appx.} about the effect of the noise level and quality factor for denoising and JPEG CAR. 

\begin{table}[!t]
    \centering
    \caption{The parameters and FLOPs comparison between SwinIR~\cite{liang2021swinir} and KGT for image denoising.}
    \label{tab:ab_efficiency2}
    \vspace{-1mm}
    \setlength{\tabcolsep}{4pt}
    \setlength{\extrarowheight}{0.5pt}
    \scalebox{0.9}{
    \begin{tabular}{cccc}
        \toprule[0.15em]
        Method & Input Size & Params (M) & FLOPs (B)\\ \hline
        SwinIR~\cite{liang2021swinir} & [1, 3, 256, 256] & 11.75 & 752.13 \\
        KGT (Ours) & [1, 3, 256, 256] & 25.85 & 134.57 \\
        \bottomrule[0.15em]
    \end{tabular}
    }
    \vspace{-1mm}
\end{table}
\begin{table*}[t]
\centering
\caption{\textit{\textbf{Grayscale image JPEG compression artifact removal}} results.} 
\label{table:jpeg_compression_artifacts_removal_gray}
\vspace{-2.5mm}
\setlength{\tabcolsep}{1.6pt}
\setlength{\extrarowheight}{0.5pt}
\scalebox{0.83}{
\begin{tabular}{c | c| c c | c c | c c | c c || c c | c c | c c | c c | c c   }
\toprule[0.15em]
\multirow{2}{*}{Set} & \multirow{2}{*}{QF} & \multicolumn{2}{c|}{JPEG}  & \multicolumn{2}{c|}{\makecell{\textcolor{magenta}{\textdagger}DnCNN3}} & \multicolumn{2}{c|}{\makecell{\textcolor{magenta}{\textdagger}DRUNet}} & \multicolumn{2}{c||}{\textcolor{magenta}{\textdagger}KGT (Ours)} & \multicolumn{2}{c|}{\makecell{GRL-S}} & \multicolumn{2}{c|}{\makecell{SwinIR}} & \multicolumn{2}{c|}{\makecell{ART}} & \multicolumn{2}{c|}{CAT}  & \multicolumn{2}{c}{KGT (Ours)}  \\ \cline{3-20}
& & PSNR & SSIM & PSNR & SSIM & PSNR & SSIM & PSNR & SSIM & PSNR & SSIM & PSNR & SSIM & PSNR & SSIM & PSNR & SSIM & PSNR & SSIM  \\
\hline
{\multirow{4}{*}{\rotatebox[origin=c]{90}{{Classic5 }}}}
	&10	&27.82	&0.7600					&29.40	&0.8030											&30.16	&0.8234		&\sotaa{30.26}	&\sotaa{0.8240}		&30.20	&\sotaa{0.8286}		&\sotab{30.27}	&0.8249		&\sotab{30.27}	&0.8258		&30.26	&0.8250					&\sotaa{30.36}	&\sotab{0.8267}		\\											
	&20	&30.12	&0.8340					&31.63	&0.8610											&32.39	&0.8734		&\sotaa{32.52}	&\sotaa{0.8740}		&32.49	&\sotaa{0.8776}		&32.52	&0.8748		&	- & - 		&32.57	&\sotab{0.8754}					&\sotaa{32.58}	&0.8748		\\											
	&30	&31.48	&0.8670					&32.91	&0.8860											&33.59	&0.8949		&\sotaa{33.74}	&\sotaa{0.8955}		&33.72	&\sotaa{0.8985}		&33.73	&0.8961		&33.74	&0.8964		&\sotaa{33.77}	&0.8964					&\sotaa{33.77}	&0.8958		\\											
	&40	&32.43	&0.8850					&33.77	&0.9000											&34.41	&0.9075		&\sotaa{34.55}	&\sotaa{0.9078}		&34.53	&\sotaa{0.9107}		&34.52	&0.9082		&34.55	&0.9086		&34.58	&\sotab{0.9087}					&\sotaa{34.57}	&0.9080		\\							
 \hline									
{\multirow{4}{*}{\rotatebox[origin=c]{90}{{LIVE1}}}}	
	&10	&27.77	&0.7730					&29.19	&0.8120											&29.79	&0.8278		&\sotaa{29.84}	&\sotaa{0.8323}		&29.82	&\sotab{0.8323}		&29.86	&0.8287		&\sotab{29.89}	&0.8300		&\sotab{29.89}	&0.8295					&\sotaa{29.92}	&\sotaa{0.8360}		\\											
	&20	&30.07	&0.8510					&31.59	&0.8800											&32.17	&0.8899		&\sotaa{32.23}	&\sotaa{0.8949}		&32.22	&\sotab{0.8930}		&32.25	&0.8909	 & - & - 		&\sotaa{32.30}	&0.8913					&\sotab{32.28}	&\sotaa{0.8950}		\\											
	&30	&31.41	&0.8850					&32.98	&0.9090											&33.59	&0.9166		&\sotaa{33.65}	&\sotaa{0.9213}		&33.65	&\sotab{0.9190}		&33.69	&0.9174		&\sotab{33.71}	&0.9178		&\sotaa{33.73}	&0.9177					&33.69	&\sotaa{0.9201}		\\											
	&40	&32.35	&0.9040					&33.96	&0.9250											&34.58	&0.9312		&\sotaa{34.65}	&\sotaa{0.9329}		&34.64	&\sotab{0.9331}		&34.67	&0.9317		&\sotab{34.70}	&0.9322		&\sotaa{34.72}	&0.9320					&34.67	&\sotaa{0.9345}\\									
\hline											
{\multirow{4}{*}{\rotatebox[origin=c]{90}{{Urban100}}}}	
	&10	&26.33	&0.7816					&28.54	&0.8484											&30.31	&0.8745		&\sotaa{30.81}	&\sotaa{0.8885}		&30.70	&0.8875		&30.55	&0.8835		&\sotab{30.87}	&\sotab{0.8894}		&30.81	&0.8866					&\sotaa{31.15}	&\sotaa{0.8941}		\\											
	&20	&28.57	&0.8545					&31.01	&0.9050											&32.81	&0.9241		&\sotaa{33.33}	&\sotaa{0.9266}		&33.24	&\sotab{0.9270}		&33.12	&0.9190		&	- & - 		&\sotab{33.38}	&0.9269					&\sotaa{33.51}	&\sotaa{0.9272}		\\											
	&30	&30.00	&0.9013					&32.47	&0.9312											&34.23	&0.9414		&\sotaa{34.74}	&\sotaa{0.9446}		&34.67	&0.9430		&34.58	&0.9417		&\sotab{34.81}	&0.9442		&\sotab{34.81}	&\sotab{0.9449}					&\sotaa{34.84}	&\sotaa{0.9462}		\\											
	&40	&31.06	&0.9215					&33.49	&0.9412											&35.20	&0.9547		&\sotaa{35.69}	&\sotaa{0.9447}		&35.62	&0.9519		&35.50	&0.9515		&35.73	&\sotaa{0.9553}		&\sotab{35.73}	&0.9511					&\sotaa{35.75}	&\sotab{0.9550}	\\				
 \bottomrule[0.15em]
\end{tabular}}
\vspace{-2mm}
\end{table*}

\begin{table*}[t]
\centering
\caption{\textit{\textbf{Color image JPEG compression artifact removal}} results.} 
\label{table:jpeg_compression_artifacts_removal_color}
\vspace{-2mm}
\setlength{\tabcolsep}{1.8pt}
\setlength{\extrarowheight}{0.5pt}
\scalebox{0.83}{
\begin{tabular}{c | c | c c | c c | c c | c c | c c || c c | c c | c c}
\toprule[0.15em]
\multirow{2}{*}{Set} & \multirow{2}{*}{QF} & \multicolumn{2}{c|}{JPEG}  & \multicolumn{2}{c|}{\makecell{\textcolor{magenta}{\textdagger}QGAC \\ \cite{ehrlich2020quantization}}} & \multicolumn{2}{c|}{\textcolor{magenta}{\textdagger}\makecell{FBCNN \\ \cite{jiang2021towards}}} & \multicolumn{2}{c|}{\textcolor{magenta}{\textdagger}\makecell{DRUNet \\ \cite{zhang2021plug}}} & \multicolumn{2}{c||}{\textcolor{magenta}{\textdagger}KGT (Ours)} & \multicolumn{2}{c|}{\makecell{SwinIR \\ \cite{liang2021swinir}}} & \multicolumn{2}{c|}{\makecell{GRL-S \\ \cite{li2023efficient}}} & \multicolumn{2}{c}{KGT (Ours)} \\ \cline{3-18}
& & PSNR & SSIM & PSNR & SSIM & PSNR & SSIM & PSNR & SSIM & PSNR & SSIM & PSNR & SSIM & PSNR & SSIM & PSNR & SSIM  \\
\hline
{\multirow{4}{*}{\rotatebox[origin=c]{90}{{LIVE1}}}}
	&10	&25.69	&0.7430		&27.62	&0.8040		&27.77	&0.8030		&27.47	&0.8045		&\sotaa{28.19}	&\sotaa{0.8146}		&28.06	&0.8129		&28.13	&0.8139		&\sotaa{28.31}	&\sotaa{0.8176}		\\
	&20	&28.06	&0.8260		&29.88	&0.8680		&30.11	&0.8680		&30.29	&0.8743		&\sotaa{30.53}	&\sotaa{0.8781}		&30.44	&0.8768		&30.49	&0.8776		&\sotaa{30.61}	&\sotaa{0.8792}		\\
	&30	&29.37	&0.8610		&31.17	&0.8960		&31.43	&0.8970		&31.64	&0.9020		&\sotaa{31.89}	&\sotaa{0.9051}		&31.81	&0.9040		&31.85	&0.9045		&\sotaa{31.94}	&\sotaa{0.9058}		\\
	&40	&30.28	&0.8820		&32.05	&0.9120		&32.34	&0.9130		&32.56	&0.9174		&\sotaa{32.81}	&\sotaa{0.9201}		&32.75	&0.9193		&32.79	&0.9195		&\sotaa{32.85}	&\sotaa{0.9204}		\\ \hline
{\multirow{4}{*}{\rotatebox[origin=c]{90}{{BSDS500}}}}
    &10	&25.84	&0.7410		&27.74	&0.8020		&27.85	&0.7990		&27.62	&0.8001		&\sotaa{28.25}	&\sotaa{0.8076}		&28.22	&0.8075		&28.26	&0.8083		&\sotaa{28.37}	&\sotaa{0.8102}		\\
	&20	&28.21	&0.8270		&30.01	&0.8690		&30.14	&0.8670		&30.39	&0.8711		&\sotaa{30.55}	&\sotaa{0.8738}		&30.54	&0.8739		&30.57	&0.8746		&\sotaa{30.63}	&\sotaa{0.8750}		\\
	&30	&29.57	&0.8650		&31.330	&0.8980		&31.45	&0.8970		&31.73	&0.9003		&\sotaa{31.90}	&\sotaa{0.9026}		&31.90	&0.9025		&31.92	&0.9030		&\sotaa{31.96}	&\sotaa{0.9035}		\\
	&40	&30.52	&0.8870		&32.25	&0.9150		&32.36	&0.9130		&32.66	&0.9168		&\sotaa{32.84}	&\sotaa{0.9190}		&32.84	&0.9189		&32.86	&0.9192		&\sotaa{32.88}	&\sotaa{0.9193}		\\

\bottomrule[0.15em]
\end{tabular}}
\vspace{-4mm}
\end{table*}
\begin{table*}[!t]
\parbox{.6\linewidth}{
\scriptsize\begin{center}
\caption{\textit{\textbf{Image Restoration in adverse weather conditions}}.}
\label{table:weather}
\vspace{-2mm}
\setlength{\extrarowheight}{3pt}
\setlength{\tabcolsep}{9pt}
\scalebox{0.9}{
\begin{tabular}{c|cc|cc|cc}
\toprule[0.15em]
    \multirow{2}{*}{\textbf{Type}} & \multicolumn{2}{c|}{\textbf{Test1 (rain+fog)}} & \multicolumn{2}{c|}{\textbf{SnowTest100k-L}} & \multicolumn{2}{c}{\textbf{RainDrop}}\\\cline{2-7}
    & Method     & PSNR        & Method  & PSNR          & Method  & PSNR             \\ \hline
    \multirow{4}{*}{\rotatebox[origin=c]{90}{\makecell{Task \\ Specific}}}
    & pix2pix~\cite{isola2017image}        & 19.09       & DesnowNet~\cite{liu2018desnownet}    & 27.17            & AttGAN~\cite{qian2018attentive}    & 30.55                 \\
    & HRGAN~\cite{li2019heavy}        & 21.56       & JSTASR~\cite{chen2020jstasr}      & 25.32            & Quan~\cite{quan2019deep}    & 31.44                  \\
    & SwinIR~\cite{liang2021swinir}       & 23.23       & SwinIR     & 28.18            & SwinIR  & 30.82                \\
    & MPRNet~\cite{zamir2021multi}        & 21.90       & DDMSNET~\cite{zhang2021deep}     & 28.85            & CCN~\cite{quan2021removing}  & 31.34                \\ \hline 
    \multirow{3}{*}{\rotatebox[origin=c]{90}{\makecell{Multi\\ Task}}}     
    & All-in-One~\cite{li2020all}     & 24.71       & All-in-One  & 28.33            & All-in-One  & \sotaa{31.12}                \\
    & TransWea.~\cite{valanarasu2022transweather}  & \sotab{27.96}       & TransWea.& \sotab{28.48}            & TransWea.       & 28.84      \\ 
    & KGT (Ours)    & \sotaa{29.57}       & KGT (Ours)& \sotaa{30.76}              & KGT (Ours)       & \sotab{30.82}   \\
    \bottomrule[0.15em]
    
\end{tabular}
}
\end{center}
}
\parbox{.4\linewidth}{
\begin{center}
\scriptsize\caption{\textit{\textbf{Image demosaicking}} results.}
\label{table:demosaicking}
\vspace{-2mm}
\setlength{\extrarowheight}{2.3pt}
\setlength{\tabcolsep}{10pt}
\scalebox{0.9}{
\begin{tabular}{l | c c}
\toprule[0.15em]
Datasets	&Kodak	&McMaster	\\ \hline
Matlab	&35.78	&34.43	\\
MMNet~\cite{kokkinos2019iterative}	&40.19	&37.09	\\
DDR~\cite{wu2016demosaicing} &41.11	&37.12	\\
DeepJoint~\cite{gharbi2016deep}	&42.00	&39.14	\\
RLDD~\cite{guo2020residual}	&42.49	&39.25	\\
DRUNet~\cite{zhang2021plug}	&42.68	&39.39	\\
RNAN~\cite{zhang2019residual}	&43.16	&39.70	\\
GRL~\cite{li2023efficient}	&\sotab{43.57}	&\sotab{40.22}	\\
KGT (Ours)	&\sotaa{43.62}	&\sotaa{40.68}	\\																			\bottomrule[0.15em]
\end{tabular}
}
\end{center}
}
\vspace{-5mm}
\end{table*}

\noindent\textbf{Efficiency Analysis.} We compare our KGT with 4 recent promising methods SwinIR, ART, CAT, and HAT-S, and HAT for x4 SR on the Urban100 dataset. The trainable parameters, the runtime, and the PSNR are reported in Tab.~\ref{tab:ab_efficiency1}. It shows that: 1) Among all the methods, HAT and KGT-S achieve 1st-class PSNR performances, reaching 28.37dB and 28.34dB, while KGT-S is much faster and has 41.7\% fewer parameters than HAT. 2) SwinIR runs a bit faster than KGT-S but with a PSNR loss of 0.89 dB. Compared with ART, HAT, and HAT-S, our KGT-S is faster and more accurate. In addition, the parameters and the FLOPs comparison between SwinIR and our KGT for denoising are reported in Tab.~\ref{tab:ab_efficiency2}. It reveals that despite SwinIR having fewer training parameters, its FLOPs significantly surpass our KGT.

\begin{figure}[!t]
    \centering
    \includegraphics[width=1.0\linewidth]{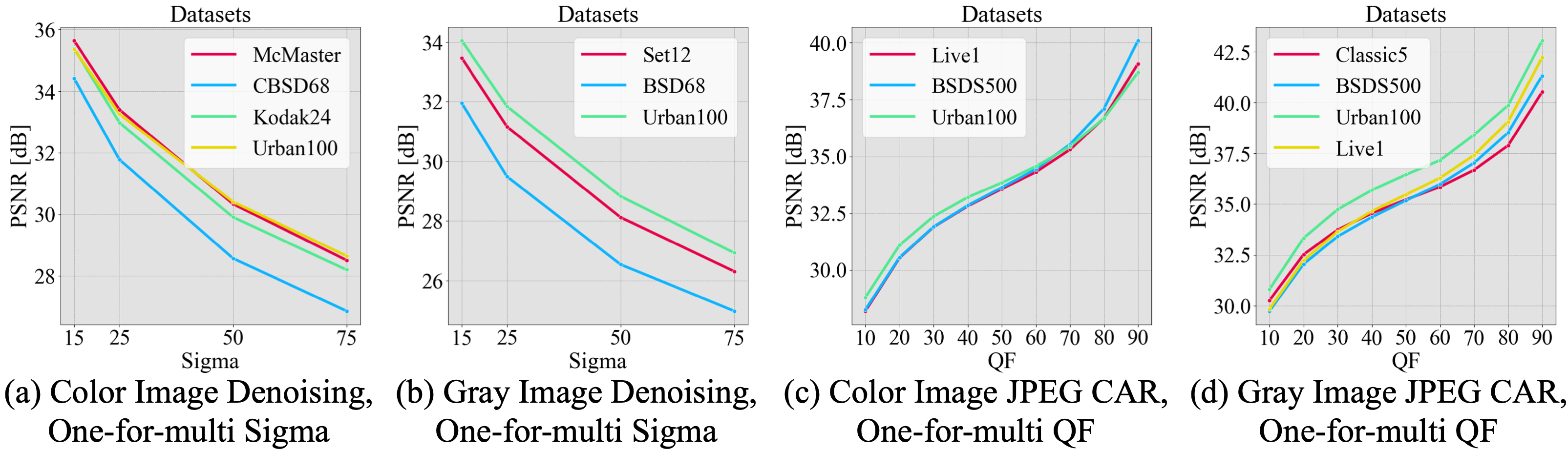}
    \vspace{-3mm}
    \caption{One model is trained to handle multiple degradation levels for denoising (a-b) and JPEG CAR (c-d).}
    \label{fig:one4m}
    \vspace{-3mm}
\end{figure}

\noindent\textbf{The Impact of One Model is Trained to Handle Multiple Degradation Levels.} Denoising and JPEG CAR are adopted for both the color and grayscale images. For denoising, Sigma is set to 15, 25, 50, and 75. For JEPG CAR, QF is set to 10, 20, 30, 40, 50, 60, 70, 80, and 90. The results in Fig.~\ref{fig:one4m} indicate that the PSNR for both tasks across all the datasets, under both color and grayscale settings, decreases when the degraded level increases. However, the proposed KGT can still outperform other comparison methods on various methods (See Tab.~\ref{table:denoising} and Tab.~\ref{table:jpeg_compression_artifacts_removal_color}). It's clear that training a single model to handle multiple degradation levels results in enhanced generalization, albeit with a slight trade-off in performance compared to its counterpart, where a distinct model is trained for each degradation level.

\subsection{Evaluation of KGT on Various IR Tasks}
\noindent\textbf{Evaluation on Image deblurring.} Tab.~\ref{table:motion_deblurring} shows the quantitative results for single image motion deblurring on synthetic datasets ({GoPro} \cite{nah2017deep}, {HIDE} \cite{shen2019human}). 
Compared to the previous state-of-the-art Restormer~\cite{zamir2022restormer}, the proposed KGT achieves significant PSNR improvement of 0.52 dB on the GoPro dataset and the second-best performance on HIDE dataset. Visual results are shown in the \textit{Appx.}.
\begin{table*}[t]
\centering
\caption{\textit{\textbf{Color and grayscale image denoising}} results. 
Both model complexity and accuracy are shown for better comparison.}
\label{table:denoising}
\vspace{-2mm}
\setlength{\extrarowheight}{0.5pt}
\setlength{\tabcolsep}{2pt}
\scalebox{0.73}{
\begin{tabular}{l | r | c c c | c c c | c c c || c c c | c c c | c c c }
\toprule[0.15em]
\multirow{3}{*}{\textbf{Method}} & \multirow{3}{*}{\# \textbf{P}} & \multicolumn{9}{c||}{\textbf{Color}} & \multicolumn{9}{c}{\textbf{Grayscale}} \\ \cline{3-20}
& & \multicolumn{3}{c|}{\textbf{CBSD68}} & \multicolumn{3}{c|}{\textbf{McMaster}} & \multicolumn{3}{c||}{\textbf{Urban100}}  & \multicolumn{3}{c|}{\textbf{Set12}} & \multicolumn{3}{c|}{\textbf{BSD68}} & \multicolumn{3}{c}{\textbf{Urban100}} \\
        &  & $\sigma$$=$$15$ & $\sigma$$=$$25$ & $\sigma$$=$$50$ & $\sigma$$=$$15$ & $\sigma$$=$$25$ & $\sigma$$=$$50$ & $\sigma$$=$$15$ & $\sigma$$=$$25$ & $\sigma$$=$$50$ & $\sigma$$=$$15$ & $\sigma$$=$$25$ & $\sigma$$=$$50$ & $\sigma$$=$$15$ & $\sigma$$=$$25$ & $\sigma$$=$$50$ & $\sigma$$=$$15$ & $\sigma$$=$$25$ & $\sigma$$=$$50$ \\ \hline
\textcolor{magenta}{\textdagger}DnCNN~\cite{kiku2016beyond}	&0.56	&33.90	&31.24	&27.95	&33.45	&31.52	&28.62	&32.98	&30.81	&27.59	&32.67	&30.35	&27.18	&31.62	&29.16	&26.23	&32.28	&29.80	&26.35	\\				
\textcolor{magenta}{\textdagger}FFDNet~\cite{zhang2018ffdnet}	&0.49	&33.87	&31.21	&27.96	&34.66	&32.35	&29.18	&33.83	&31.40	&28.05	&32.75	&30.43	&27.32	&31.63	&29.19	&26.29	&32.40	&29.90	&26.50	\\				
\textcolor{magenta}{\textdagger}IRCNN	&0.19	&33.86	&31.16	&27.86	&34.58	&32.18	&28.91	&33.78	&31.20	&27.70	&32.76	&30.37	&27.12	&31.63	&29.15	&26.19	&32.46	&29.80	&26.22	\\				
\textcolor{magenta}{\textdagger}DRUNet~\cite{zhang2021plug}	&32.64	&34.30	&\sotab{31.69}	&28.51	&35.40	&33.14	&30.08	&34.81	&32.60	&29.61	&33.25	&30.94	&27.90	&\sotab{31.91}	&29.48	&26.59	&33.44	&31.11	&27.96	\\				
\textcolor{magenta}{\textdagger}Restormer~\cite{zamir2022restormer}	&26.13	&\sotab{34.39}	&\sotaa{31.78}	&\sotaa{28.59}	&\sotab{35.55}	&\sotab{33.31}	&\sotab{30.29}	&\sotab{35.06}	&\sotab{32.91}	&\sotab{30.02}	&\sotab{33.35}	&\sotab{31.04}	&\sotab{28.01}	&\sotaa{31.95}	&\sotaa{29.51}	&\sotaa{26.62}	&\sotab{33.67}	&\sotab{31.39}	&\sotab{28.33}	\\				
\textcolor{magenta}{\textdagger}KGT (Ours)	&25.82	&\sotaa{34.42}	&\sotaa{31.78}	&\sotab{28.57}	&\sotaa{35.65}	&\sotaa{33.40}	&\sotaa{30.34}	&\sotaa{35.37}	&\sotaa{33.26}	&\sotaa{30.41}	&\sotaa{33.47}	&\sotaa{31.16}	&\sotaa{28.12}	&\sotaa{31.95}	&\sotab{29.49}	&\sotab{26.54}	&\sotaa{34.05}	&\sotaa{31.84}	&\sotaa{28.83}	\\				
\hline																				
DnCNN~\cite{kiku2016beyond}	&0.56	&33.90	&31.24	&27.95	&33.45	&31.52	&28.62	&32.98	&30.81	&27.59	&32.86	&30.44	&27.18	&31.73	&29.23	&26.23	&32.64	&29.95	&26.26	\\				
RNAN~\cite{zhang2019residual}	&8.96	&-	&-	&28.27	&-	&-	&29.72	&-	&-	&29.08	&-	&-	&27.70	&-	&-	&26.48	&-	&-	&27.65	\\				
IPT~\cite{chen2021pre}	&115.33	&-	&-	&28.39	&-	&-	&29.98	&-	&-	&29.71	&-	&-	&-	&-	&-	&-	&-	&-	&-	\\				
EDT-B~\cite{li2021efficient}	&11.48	&34.39	&31.76	&28.56	&35.61	&33.34	&30.25	&35.22	&33.07	&30.16	&-	&-	&-	&-	&-	&-	&-	&-	&-	\\				
DRUNet~\cite{zhang2021plug}	&32.64	&34.30	&31.69	&28.51	&35.40	&33.14	&30.08	&34.81	&32.60	&29.61	&33.25	&30.94	&27.90	&31.91	&29.48	&26.59	&33.44	&31.11	&27.96	\\				
SwinIR~\cite{liang2021swinir}	&11.75	&\sotab{34.42}	&31.78	&28.56	&35.61	&33.20	&30.22	&35.13	&32.90	&29.82	&33.36	&31.01	&27.91	&\sotab{31.97}	&29.50	&26.58	&33.70	&31.30	&27.98	\\				
Restormer~\cite{zamir2022restormer}	&26.13	&34.40	&31.79	&\sotab{28.60}	&35.61	&33.34	&30.30	&35.13	&32.96	&30.02	&33.42	&31.08	&28.00	&31.96	&\sotab{29.52}	&26.62	&33.79	&31.46	&28.29	\\				
Xformer~\cite{zhang2023xformer}	& 25.23	&\sotaa{34.43}	&\sotaa{31.82}	&\sotaa{28.63}	&\sotaa{35.68}	&\sotaa{33.44}	&\sotab{30.38}	&\sotab{35.29}	&\sotab{33.21}	&\sotab{30.36}	&\sotab{33.46}	&\sotab{31.16}	&\sotab{28.10}	&\sotaa{31.98}	&\sotaa{29.55}	&\sotaa{26.65}	&\sotab{33.98}	&\sotab{31.78}	&\sotab{28.71}	\\								
KGT (Ours)	& 25.82	&\sotaa{34.43}	&\sotab{31.79}	&\sotab{28.60}	&\sotab{35.65}	&\sotab{33.43}	&\sotaa{30.38}	&\sotaa{35.38}	&\sotaa{33.29}	&\sotaa{30.51}	&\sotaa{33.48}	&\sotaa{31.18}	&\sotaa{28.14}	&\sotab{31.97}	&\sotab{29.52}	&\sotab{26.53}	&\sotaa{34.09}	&\sotaa{31.87}	&\sotaa{28.86}	\\				
\bottomrule[0.15em]
\end{tabular}}
\vspace{-1mm}
\end{table*}

\noindent\textbf{Evaluation on JPEG CAR.} The experiments for both grayscale and color images are conducted with 4 image quality factors ranging from 10 to 40 under two settings (\ie \textcolor{magenta}{\textdagger} a single model is trained to handle multiple quality factors, and each model for each quality). 
The quantitative results for color images in \textit{Appx.} show that our KGT achieves the best results on all the test sets and quality factors among all compared methods like QGAC, FBCNN, DRUNet, SwinIR, and GRL-S. For grayscale, the results are shown in \textit{Appx.} also validate that our KGT outperforms all other methods like DnCNN-3, DRUNet, GRL-S, SwinIR, ART, and CAT under both settings. The visual comparisons in the \textit{Appx.} further supports the effectiveness of our method.

\noindent\textbf{Evaluation on Image Denoising.} We show color and grayscale image denoising results in Tab.~\ref{table:denoising} under two settings (\emph{i.e.}, \textcolor{magenta}{\textdagger} one model for all noise levels $\sigma = \{15, 25, 50\}$ and each model for each noise level). 
For a fair comparison, both the parameter and accuracy are reported for all the methods.
For \textcolor{magenta}{\textdagger}, our KGT performs better on all test sets for color and grayscale image denoising compared to others. 
It's worth noting that we outperform DRUNet and Restormer with lower trainable parameters. For another setting, our KGT also archives better results on CBSD68 and Urban100 for color image denoising, and on Set12 and Urban100 for grayscale denoising. 
These interesting comparisons validate the effectiveness of the proposed KGT and also indicate that KGT has a higher generalization ability. The visual results in \textit{Appx.} also support that the proposed KGT can remove heavy noise corruption and preserve high-frequency image details, resulting in sharper edges and more natural textures without over-smoothness or over-sharpness problems.

\begin{table*}[t]
\scriptsize
\setlength{\tabcolsep}{7pt}
\setlength{\extrarowheight}{0.9pt}

\setlength{\abovecaptionskip}{0.1cm}
\begin{center}
\caption{\textbf{\textit{Classical image SR}} results. Both lightweight and accurate models are summarized.}
\label{tab:sr_results}
\begin{tabular}{l|c|r|cc|cc|cc|cc|cc}
\toprule[0.15em]
\multirow{2}{*}{\textbf{Method}} & \multirow{2}{*}{\textbf{Scale}} & {\textbf{Params}} &  \multicolumn{2}{c|}{\textbf{Set5}} &  \multicolumn{2}{c|}{\textbf{Set14}} &  \multicolumn{2}{c|}{\textbf{BSD100}} &  \multicolumn{2}{c|}{\textbf{Urban100}} &  \multicolumn{2}{c}{\textbf{Manga109}} 
\\
\cline{4-13}
&  & \multicolumn{1}{c|}{\textbf{[M]}} & PSNR & SSIM & PSNR & SSIM & PSNR & SSIM & PSNR & SSIM & PSNR & SSIM
\\
\hline
RCAN~\cite{zhang2018rcan}&	$\times$2&	15.44&	38.27&	0.9614&	34.12&	0.9216&	32.41&	0.9027&	33.34&	0.9384&	39.44&	0.9786\\
SAN~\cite{dai2019SAN}&	$\times$2&	15.71&	38.31&	0.9620&	34.07&	0.9213&	32.42&	0.9028&	33.10&	0.9370&	39.32&	0.9792\\
HAN~\cite{niu2020HAN}&	$\times$2&	63.61&	38.27&	0.9614&	34.16&	0.9217&	32.41&	0.9027&	33.35&	0.9385&	39.46&	0.9785\\
IPT~\cite{chen2021pre}&	$\times$2&	115.48&	38.37&	-&	34.43&	-&	32.48&	-&	33.76&	-&	-&	-\\ \hline
SwinIR~\cite{liang2021swinir}&	$\times$2&	11.75&	38.42&	0.9623&	34.46&	0.9250&	32.53&	0.9041&	33.81&	0.9427&	39.92&	0.9797\\  
CAT-A~\citep{chen2022cross}    & $\times$2 &16.46 & 38.51 & 0.9626 & 34.78 & 0.9265 & 32.59 & 0.9047 & 34.26 & 0.9440 & 40.10 & 0.9805 \\
ART~\cite{zhang2023accurate}	&$\times$2 &16.40	&38.56	&0.9629	&34.59	&0.9267	&32.58	&0.9048	&34.30	&0.9452	&40.24	&0.9808	\\		
EDT~\cite{li2021efficient}&	$\times$2&	11.48&	 \sotaa{38.63}&	 {0.9632}&	 {34.80}&	0.9273&	 {32.62}&	0.9052&	34.27&	\sotab{0.9456}&	 {40.37}&	 {0.9811}\\ 

KGT-S (Ours)	&$\times$2& 11.87	&38.57	&\sotab{0.9651}	&\sotab{34.99}	&\sotab{0.9300}	&\sotab{32.65}	&\sotab{0.9078}	&\sotab{34.86}	&\sotaa{0.9472}	&\sotab{40.45}	&\sotab{0.9824}			\\																			
KGT-B (Ours) &	$\times$2&	19.90&	 \sotab{38.61}&	 \sotaa{0.9654}&	 \sotaa{35.08}&	 \sotaa{0.9304}&	 \sotaa{32.69}&	 \sotaa{0.9084}&	 \sotaa{34.99}&	 
{0.9455}&	 \sotaa{40.59}&	 \sotaa{0.9830}\\ 

\midrule[0.1em]

RCAN~\cite{zhang2018rcan}&	$\times$3&	15.63&	34.74&	0.9299&	30.65&	0.8482&	29.32&	0.8111&	29.09&	0.8702&	34.44&	0.9499\\
SAN~\cite{dai2019SAN}&	$\times$3&	15.90&	34.75&	0.9300&	30.59&	0.8476&	29.33&	0.8112&	28.93&	0.8671&	34.30&	0.9494\\
HAN~\cite{niu2020HAN}&	$\times$3&	64.35&	34.75&	0.9299&	30.67&	0.8483&	29.32&	0.8110&	29.10&	0.8705&	34.48&	0.9500\\
NLSA~\cite{mei2021NLSA}&	$\times$3&	45.58&	34.85&	0.9306&	30.70&	0.8485&	29.34&	0.8117&	29.25&	0.8726&	34.57&	0.9508\\
IPT~\cite{chen2021pre}&	$\times$3&	115.67&	34.81&	-&	30.85&	-&	29.38&	-&	29.49&	-&	-&	-\\ \hline
SwinIR~\cite{liang2021swinir}&	$\times$3&	11.94&	34.97&	0.9318&	30.93&	0.8534&	29.46&	0.8145&	29.75&	0.8826&	35.12&	0.9537\\  
CAT-A~\citep{chen2022cross}      & $\times$3 &16.64 & 35.06 & 0.9326 & 31.04 & 0.8538 & 29.52 & 0.8160 & 30.12 & 0.8862 & 35.38 & 0.9546 \\
ART~\cite{zhang2023accurate}		&$\times$3 &16.58	&35.07	&0.9325	&31.02	&0.8541	&29.51	&0.8159	&30.10	&0.8871	&35.39	&0.9548	\\		
EDT~\cite{li2021efficient}&	$\times$3&	11.66&	\sotaa{35.13}&	0.9328&	31.09&	0.8553&	29.53&	0.8165&	30.07&	0.8863&	35.47&	0.9550\\ 
KGT-S (Ours)	&$\times$3&	12.05 &34.99	&\sotab{0.9366}	&\sotab{31.23}	&\sotab{0.8594}	&\sotab{29.53}	&\sotab{0.8223}	&\sotab{30.71}	&\sotab{0.8950}	&\sotab{35.52}	&\sotab{0.9573}			\\														
KGT-B (Ours) &	$\times$3&	20.08&	\sotab{35.03}&	\sotaa{0.9371}&	\sotaa{31.29}&	\sotaa{0.8603}&	\sotaa{29.54}&	\sotaa{0.8227}&	\sotaa{30.87}&	\sotaa{0.9012}&	\sotaa{35.60}&	\sotaa{0.9581}	\\ 
\midrule[0.1em]

RCAN~\cite{zhang2018rcan}&	$\times$4&	15.59&	32.63&	0.9002&	28.87&	0.7889&	27.77&	0.7436&	26.82&	0.8087&	31.22&	0.9173\\
SAN~\cite{dai2019SAN}&	$\times$4&	15.86&	32.64&	0.9003&	28.92&	0.7888&	27.78&	0.7436&	26.79&	0.8068&	31.18&	0.9169\\
HAN~\cite{niu2020HAN}&	$\times$4&	64.20&	32.64&	0.9002&	28.90&	0.7890&	27.80&	0.7442&	26.85&	0.8094&	31.42&	0.9177\\
IPT~\cite{chen2021pre}&	$\times$4&	115.63&	32.64&	-&	29.01&	-&	27.82&	-&	27.26&	-&	-&	-\\ \hline
SwinIR~\cite{liang2021swinir}&	$\times$4&	11.90&	32.92&	0.9044&	29.09&	0.7950&	27.92&	0.7489&	27.45&	0.8254&	32.03&	0.9260\\  
CAT-A~\citep{chen2022cross}      & $\times$4 &16.60 & {33.08} & 0.9052 & 29.18 & 0.7960 & 27.99 & 0.7510 & 27.89 & 0.8339 & 32.39 & 0.9285 \\
ART~\cite{zhang2023accurate}	&$\times$4	& 16.55 &33.04	&0.9051	&29.16	&0.7958	&27.97	&0.751	&27.77	&0.8321	&32.31	&0.9283	\\		
EDT~\cite{li2021efficient}&	$\times$4&	11.63&	\sotab{33.06}&	0.9055&	 {29.23}&	0.7971&	 \sotaa{27.99}&	0.7510&	27.75&	\sotab{0.8317}&	 {32.39}&	 {0.9283}\\ 

KGT-S (Ours)	&$\times$4& 12.02	&33.02	&\sotab{0.9082}	&\sotab{29.29}	&\sotab{0.8026}	&27.96	&\sotab{0.7582}	&\sotab{28.34}	&\sotaa{0.8467}	&\sotab{32.48}	&\sotab{0.9322}			\\				
KGT-B (Ours) &	$\times$4&	20.04&	 \sotaa{33.08}&	 \sotaa{0.9090}&	 \sotaa{29.34}&	 \sotaa{0.8037}&	 \sotab{27.98}&	 \sotaa{0.7599}&	 \sotaa{28.51}&	 \sotaa{0.8467}&	 \sotaa{32.56}&	 \sotaa{0.9335}\\ 
\bottomrule[0.15em]
\end{tabular}
\end{center}
\vspace{-5mm}
\end{table*}
\begin{figure*}[!t]
    \centering
    \includegraphics[width=0.99\linewidth]{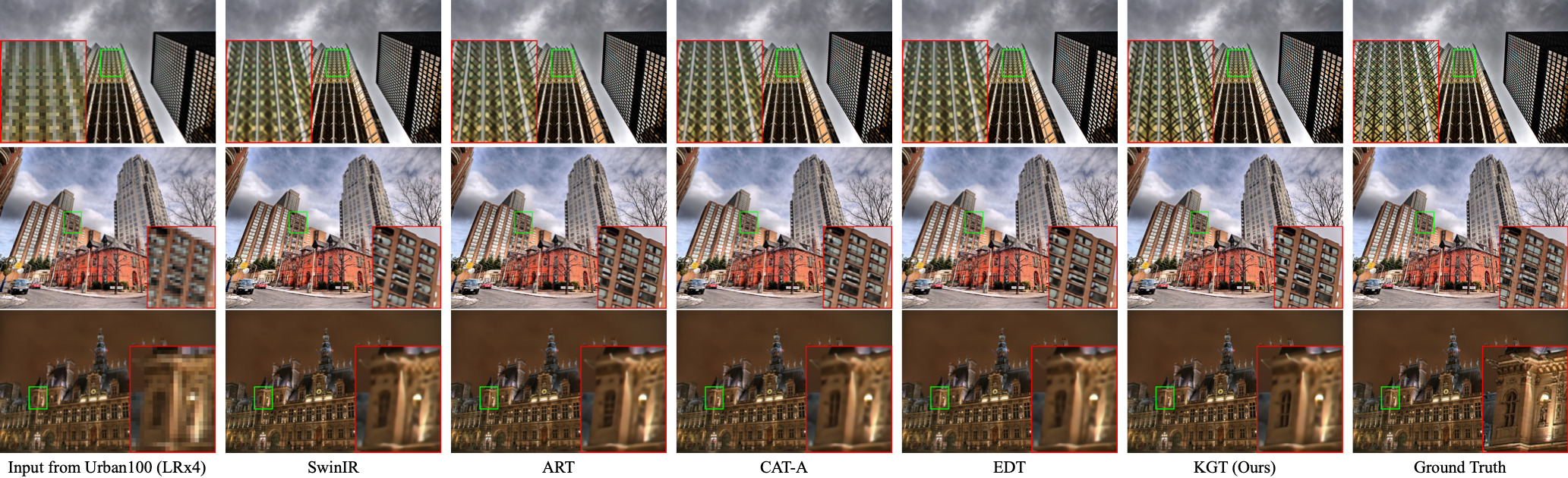}
    \vspace{-2mm}
    \caption{Visual comparison of classical image SR (x4) on Urban100. Best viewed by zooming.}
    \label{fig:sr_x4}
    \vspace{-3mm}
\end{figure*}

\noindent\textbf{Evaluation in AWC.}
We validate KGT in adverse weather conditions like rain+fog (Test1), snow (SnowTest100K), and raindrops (RainDrop). 
PSNR is reported in Tab.~\ref{table:weather}. Our method achieves the best performance on Test1 (\emph{i.e.}, 5.76\% improvement) and SnowTest100k-L (\emph{i.e.} 8.01\% improvement), while the second-best PSNR on RainDrop compared to all other methods. The visual comparison is in our \textit{Appx.}. 

\noindent\textbf{Evaluation on Image Demosaicking.} The quantitative results shown in \ref{table:demosaicking} indicate that the proposed KGT archives the best performance on both the Kodak and MaMaster test sets. Especially, 0.05dB and 0.45dB absolute improvement compared to the current state-of-the-art. 

\noindent\textbf{Evaluation on SR.} For the classical image SR, we compared our KGT with both recent lightweight and accurate SR models, and the quantitative results are shown in Tab.~\ref{tab:sr_results}. Compared to EDT, KGT-base achieves significant improvements on Urban100 (\emph{i.e.}, 0.72 dB and 0.76dB for x2 and x4 SR) and Manga109 datasets (\emph{i.e.}, 0.22dB and 0.17 dB for x2 and x4 SR). Furthermore, even the KGT-small consistently ranks as the runner-up in terms of performance across the majority of test datasets, all while maintaining a reduced number of trainable parameters. The visual results shown in Fig.~\ref{fig:sr_x4} also validate the effectiveness of the proposed KGT in restoring more details and structural content.

\section{Conclusion}
\label{sec:conclusion}
In this paper, for the first time, we utilize ViTs from the graph perspective specifically tailored for IR with the proposed KGT for both the widely-used multi-stage (For image SR) and the U-shaped architectures (For other IR tasks). In particular, a Key-Graph is constructed that can capture the complex relation of each node feature with only the most relevant topk nodes with the proposed Key-Graph constructor instead of a dense fully connected graph. Then the Key-Graph is shared with all the KGT layers within the same stage, which enables the Key-Graph Attention to capture fewer but the key relation of each node.
As a result, KGT leads to the window-wise computation complexity reduced from $\mathcal{O}((hw)^{2})$ to $\mathcal{O}((hw) \times k)$, which largely released the potential of ViTs in a sparse yet representative manner. Extensive experiments on 6 IR tasks validated the effectiveness of the proposed KGT and the results demonstrate that the proposed KGT achieves new state-of-the-art performance. Our KGT reveals that the global cues are essential, but not all the global cues are necessary. The code will be available.

{
    \small
    \bibliographystyle{ieeenat_fullname}
    \bibliography{main}
}

\end{document}